\documentclass[10pt,journal]{IEEEtran}

\usepackage[T1]{fontenc}
\usepackage[utf8]{inputenc}
\usepackage{cite}
\usepackage{amsmath,amssymb}
\usepackage{booktabs}
\usepackage{tabularx}
\usepackage{array}
\usepackage{multirow}
\usepackage{xcolor}
\usepackage{tikz}
\usetikzlibrary{arrows.meta,positioning,fit,shapes.geometric}
\usepackage[hidelinks]{hyperref}
\usepackage{placeins}

\usepackage{algorithm}
\usepackage{algorithmicx}
\usepackage{algpseudocode}

\newcolumntype{Y}{>{\raggedright\arraybackslash}X}

\title{Planning-Oriented End-to-End Autonomous Driving: Architectures, Evaluation, and Emerging Paradigms}

\author{Yanchen~Guan,
        Xingcheng~Liu,
        Bin~Rao,
        Chengyue~Wang,
        Guofa~Li, 
        Yunjian~Li,
        Lishengsa~Yue,
        Zhiyong~Cui,
       Chengzhong~Xu,~\IEEEmembership{Fellow,~IEEE}
       and~Zhenning~Li$^{*}$% <-this % stops a space
\thanks{*\,Corresponding author. E-mail: zhenningli@um.edu.mo}% <-this % stops a space
\thanks{Y. Guan, X. Liu, B. Rao, C. Wang, C. Xu and Z. Li are with the State Key Laboratory of Internet of Things for Smart City, University of Macau, Macau SAR, 999078, China}
\thanks{G. Li is with the College of Mechanical and Vehicle Engineering, Chongqing University, Chongqing, 400044, China}
\thanks{Y. Li is with the Department of Civil and Environmental Engineering, The Hong Kong University of Science and Technology, Hong Kong, 999077, China}
\thanks{L. Yue is with the School of Transportation Engineering, Tongji University, No. 4800 Cao’an Road, Shanghai, 201804, China}
\thanks{Z. Cui is with the School of Transportation Science and Engineering, School of Software Engineering, Beihang University, Beijing, 100191, China}}

\begin{document}
\maketitle

\begin{abstract}
End-to-end autonomous driving has evolved from camera-to-control regression toward planning-oriented systems that use structured representations, trajectory-level outputs, and increasingly realistic evaluation protocols. This survey reviews this transition across behavior cloning, conditional imitation learning, privileged distillation, BEV and vectorized planning, unified perception-prediction-planning architectures, world-model-based planners, and vision-language-action systems. We argue that the key distinction in modern end-to-end driving is not whether intermediate representations are used, but whether they are learned, supervised, and evaluated to support safe, feasible, and route-compliant planning. To organize the literature, we synthesize existing methods along four axes: input representation, planning output, supervision signal, and evaluation protocol. We further examine the benchmark shift from open-loop trajectory matching to closed-loop simulation, non-reactive real-log evaluation, long-tail testing, and human-preference-aware metrics. Our analysis highlights that architectural progress is difficult to interpret without benchmark-consistent evaluation, and that displacement-based open-loop metrics alone provide limited evidence for safe and human-aligned driving. We conclude with open challenges in uncertainty-aware planning, learner-expert mismatch, runtime safety assurance, language-action grounding, world-model validation, and reproducible benchmarking.
\end{abstract}

\begin{IEEEkeywords}
End-to-end autonomous driving, trajectory planning, closed-loop evaluation, world models, vision-language-action models.
\end{IEEEkeywords}

\section{Introduction}
\IEEEPARstart{A}{utonomous} driving systems have traditionally been engineered as modular pipelines in which perception, prediction, planning, and control are designed as separate subsystems \cite{paden2016survey,gonzalez2015review,katrakazas2015real}. This decomposition is attractive because each module has a clear technical objective and a relatively interpretable failure surface. However, it also creates brittle interfaces across the driving stack. Perception modules may optimize detection metrics that are only weakly related to planning, prediction modules may produce futures that downstream planners cannot effectively use, and planning modules may operate on delayed or discretized abstractions that discard information contained in the original sensor stream \cite{karle2022scenario}. End-to-end autonomous driving (E2E-AD) emerged as a response to this interface problem. Its central goal is to learn a task-level mapping from observations, ego state, maps, route intent, and, in some recent systems, language instructions to driving controls, waypoints, or motion plans optimized for the final driving task \cite{bojarski2016end,codevilla2018end,chen2024end}.

The term ``end-to-end,'' however, has become increasingly overloaded. In its narrowest form, it refers to direct sensor-to-control regression, as in early camera-to-steering systems \cite{bojarski2016end}. Contemporary E2E-AD systems are often very different from this early formulation. Many influential methods retain structured intermediate representations, including BEV features, vectorized scene tokens, occupancy fields, object queries, world-model latents, route encodings, language tokens, and safety filters \cite{hu2023planning,jiang2023vad,jia2025drivetransformer,zheng2025world4drive}. These systems are not end-to-end because they remove structure. Rather, they are end-to-end in the sense that their learned representations, training objectives, and planning heads are coupled around the final driving behavior. This survey therefore adopts the term \emph{planning-oriented end-to-end driving} to describe learned driving systems whose architectures and supervision signals are organized around producing safe, feasible, and route-compliant plans or controls.

This planning-oriented view is timely for three reasons. First, the field has moved beyond the question of whether imitation learning can make a vehicle follow a lane. Current research increasingly focuses on how learned driving policies behave over long horizons, under distribution shift, in interactive traffic, and in rare but safety-critical scenarios. Second, benchmark design has become a central methodological issue. Open-loop displacement errors on logged trajectories are scalable and easy to reproduce, but they often fail to measure route progress, collision avoidance, rule compliance, comfort, or recovery from the policy's own mistakes \cite{li2024ego,jaeger2023hidden}. Benchmarks such as nuPlan, Bench2Drive, NAVSIM, and WOD-E2E have therefore been introduced to evaluate planning quality under more realistic or planning-aware protocols \cite{caesar2021nuplan,jia2024bench2drive,dauner2024navsim,xu2026wod}. Third, world models and vision-language-action (VLA) systems are expanding the meaning of planning itself. Modern systems may predict latent future states, evaluate candidate trajectories through imagined rollouts, align motion outputs with language supervision, or use language-based reasoning to support action generation \cite{li2025end,zheng2025world4drive,renz2025simlingo,fu2025orion,xu2026wam,yang2026worldrft}.

This survey makes four contributions. First, it provides a planning-oriented taxonomy of E2E-AD along four axes: input representation, planning output, supervision signal, and evaluation protocol. Second, it reviews the historical transition from behavior cloning and conditional imitation learning to structured planning architectures, world-model-based planners, and foundation-model-based driving systems. Third, it analyzes the benchmark shift from open-loop trajectory matching to closed-loop simulation, non-reactive real-log evaluation, long-tail testing, and human-preference-aware metrics. Fourth, it identifies open research problems in uncertainty-aware planning, learner-expert mismatch, runtime safety assurance, language-action grounding, world-model validation, and reproducible benchmarking. The central thesis of this survey is that planning-oriented E2E-AD should not be judged by how structure-free it appears architecturally, but by whether its learned structure, supervision, and evaluation protocol jointly support safe, reliable, and reproducible planning.

\section{Scope, Methodology, and Related Surveys}

\subsection{Motivation and Positioning}
Recent surveys have reviewed end-to-end driving, vision-language driving, world models, foundation models, and autonomous-driving datasets from different perspectives \cite{claussmann2019review,chen2024end,chib2023recent,zhou2024vision,jiang2025survey,guan2024world,feng2025survey,gao2025survey}. This survey is motivated not by the absence of prior summaries, but by the need to connect three issues that are increasingly inseparable in planning-oriented E2E-AD: model architecture, evaluation protocol, and emerging foundation-model-based driving paradigms.

In current E2E-AD research, a method cannot be assessed only by its backbone, sensor modality, or intermediate representation. It is also necessary to examine its planning output, supervision signal, evaluation protocol, use of language or world modeling, and transferability from open-loop metrics to closed-loop, long-tail, or preference-aware evaluation. This survey therefore treats planning as the organizing endpoint of E2E-AD and examines how perception, prediction, language reasoning, world modeling, and benchmarking contribute to safe, feasible, and route-compliant ego plans. Table~\ref{tab:related_surveys} summarizes the distinction between this survey and existing surveys.

\begin{table*}[t]
\caption{Related Surveys and This Survey's Positioning.}
\label{tab:related_surveys}
\centering
\setlength{\extrarowheight}{2pt}
\begin{tabularx}{\textwidth}{p{0.20\textwidth}p{0.20\textwidth}Y Y}
\toprule
Existing survey line & Typical scope & Less emphasized issue & This survey's angle \\
\midrule
Broad E2E-AD surveys \cite{chen2024end,chib2023recent} & General E2E driving motivations, architectures, datasets, and challenges. & The post-2024 benchmark mismatch around NAVSIM, Bench2Drive, WOD-E2E, and cross-protocol correlation. & Planning-oriented taxonomy centered on output spaces, supervision, and evaluation validity. \\
Vision-language driving surveys \cite{zhou2024vision,jiang2025survey} & VLM/VLA perception, reasoning, instruction following, and data generation. & Whether language improves closed-loop planning rather than only QA or explanation. & Action-grounded VLA analysis: language as annotation, instruction, teacher, or action generator. \\
World-model surveys \cite{guan2024world,feng2025survey} & Generative prediction, scene evolution, simulation, and latent dynamics. & The difference between realistic generation and safe trajectory selection. & Consequence-aware planning view: representation pretraining, future prediction, trajectory evaluation, and reactive simulation. \\
Foundation-model surveys for AD \cite{gao2025survey} & Large models for perception, simulation, reasoning, and decision making. & Planning-specific benchmark evidence and deployment constraints. & Connects foundation-model roles to trajectory output, latency, safety envelopes, and reproducibility. \\
Benchmark and dataset surveys \cite{liu2024survey,song2023synthetic} & Dataset scale, sensor modalities, annotation types, and task definitions. & Method taxonomy and how metric choices change E2E claims. & Benchmark-consistent comparison of open-loop, non-reactive, closed-loop, long-tail, and preference-aware evaluation. \\
\bottomrule
\end{tabularx}
\end{table*}

\subsection{Review Protocol}
\begin{figure*}[t]
\centering
\includegraphics[width=0.90\textwidth]{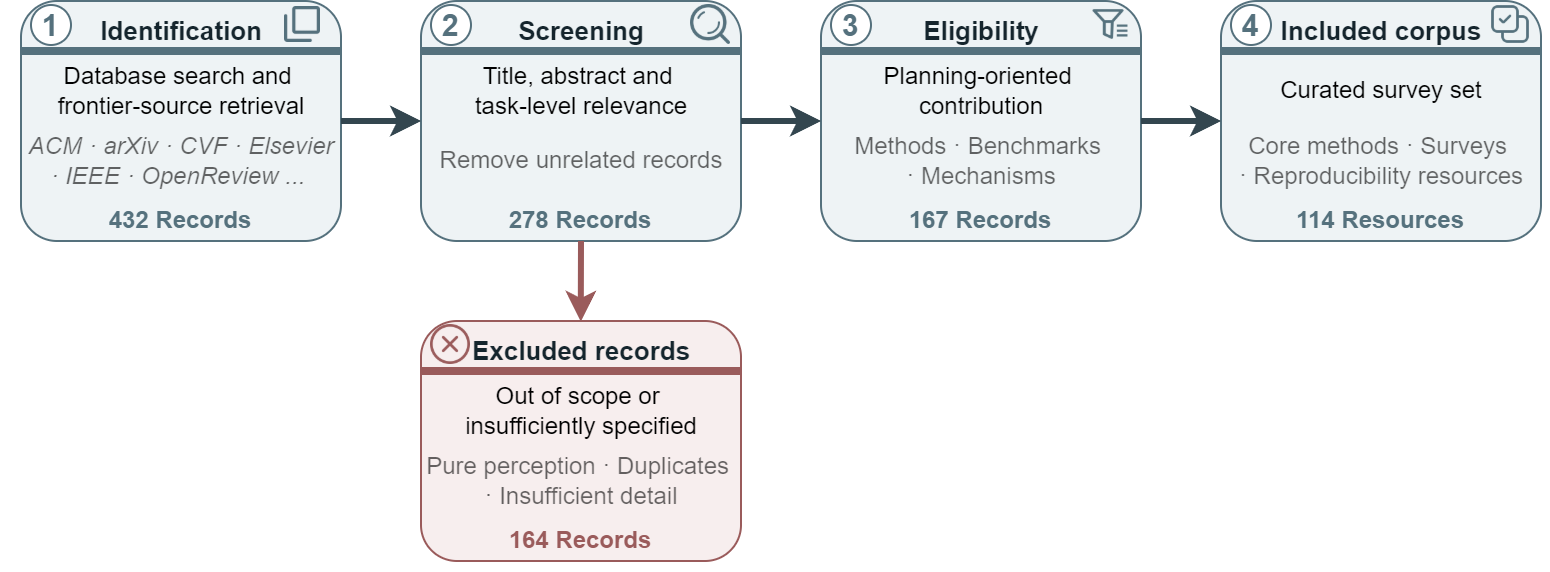}
\caption{Search-and-screen workflow used for this survey. The workflow records the main sources, search terms, citation tracing, screening criteria, and final categorization of the reviewed literature.}
\label{fig:search_flow}
\end{figure*}

We adopted a structured narrative review protocol to cover both established E2E-AD literature and rapidly emerging work across peer-reviewed venues, arXiv preprints, benchmark repositories, challenge leaderboards, code releases, and official project pages. As shown in Figure \ref{fig:search_flow}, the primary sources included IEEE Xplore, ACM Digital Library, SpringerLink, ScienceDirect/Elsevier, arXiv, OpenReview, CVF Open Access, NeurIPS proceedings, public benchmark repositories, and official project websites.

The main search period covered work from 2015 to June 2026, with earlier studies included when they shaped neural or end-to-end driving. The core search terms included ''end-to-end autonomous driving,'' learning-based planning,'' closed-loop driving,'' autonomous driving planning benchmark,'' nuPlan,'' NAVSIM,'' Bench2Drive,'' WOD-E2E,'' world model autonomous driving,'' vision-language driving,'' VLA driving,'' long-tail autonomous driving,'' and ``autonomous driving generalization benchmark.'' We also conducted backward and forward citation tracing from highly connected papers and benchmarks, including PilotNet, Conditional Imitation Learning, Learning by Cheating, TransFuser, TCP, UniAD, VAD, NAVSIM, Bench2Drive, DriveLM, SimLingo, WOD-E2E, and LEAD.

\subsection{Inclusion and Exclusion Criteria}
A work was included if it satisfied at least one criterion: 1) introducing an influential E2E driving formulation; 2) predicting planning-relevant outputs such as controls, waypoints, trajectories, trajectory distributions, or action tokens; 3) proposing a benchmark, dataset, simulator, or metric for planning evaluation; 4) contributing a world-model, VLM, or VLA mechanism relevant to ego planning; 5) releasing a public dataset, benchmark, codebase, evaluation server, or reproducibility resource; or 6) providing a safety, interpretability, robustness, or evaluation critique of E2E driving.

We excluded studies focused only on perception, detection, segmentation, tracking, or motion prediction unless they directly supported ego planning or were widely used in an E2E driving stack. Purely proprietary systems were also excluded when technical details were insufficient for comparison. For each retained work, we recorded its input representation, planning output, supervision signal, evaluation protocol, benchmark evidence, and public-resource status when available.

\subsection{Scope and Definitions}
\subsubsection{Definition of End-to-End Driving}
An autonomous driving stack can be called end-to-end when the mapping from driving inputs to driving outputs is optimized as a learned policy or differentiable system whose final objective is a driving action, route-conditioned trajectory, or motion plan. Inputs may include camera images, LiDAR, radar, ego state, HD maps, route commands, goal points, or natural language \cite{deng2023v2x}. Outputs may include steering and throttle, waypoints, trajectories, trajectory distributions, or discrete action tokens.

This definition intentionally includes systems with structured intermediate representations. DeepDriving predicted affordances rather than direct control \cite{chen2015deepdriving}; TCP coupled trajectory and control prediction \cite{wu2022trajectory}; UniAD organized perception, prediction, occupancy, mapping, and planning around the final planning target \cite{hu2023planning}; VAD introduced vectorized scene representations \cite{jiang2023vad}; DriveTransformer used sparse task queries for scalable E2E learning \cite{jia2025drivetransformer}; and World4Drive used latent world representations for perception-annotation-free planning \cite{zheng2025world4drive}. These systems are end-to-end not because they are structureless, but because their structures are learned, optimized, or coupled through the planning objective.

\subsubsection{Planning-Oriented Versus Generic End-to-End}
A generic E2E survey can be organized around sensors, neural backbones, or chronological method families. A planning-oriented survey instead asks how each design choice affects the generation, evaluation, and safety of future ego motion. In this view, perception is relevant when it supports planning; a world model is relevant when it predicts the consequences of candidate actions; a VLM is relevant when its reasoning is grounded in the action space; and a benchmark is relevant when its metrics provide evidence for safe and reliable driving behavior. We therefore treat planning as the conceptual endpoint of the literature rather than as one module among many.

% \FloatBarrier
\section{Evolution Toward Planning-Oriented E2E-AD}
\begin{figure*}[t]
\centering
\includegraphics[width=0.90\textwidth]{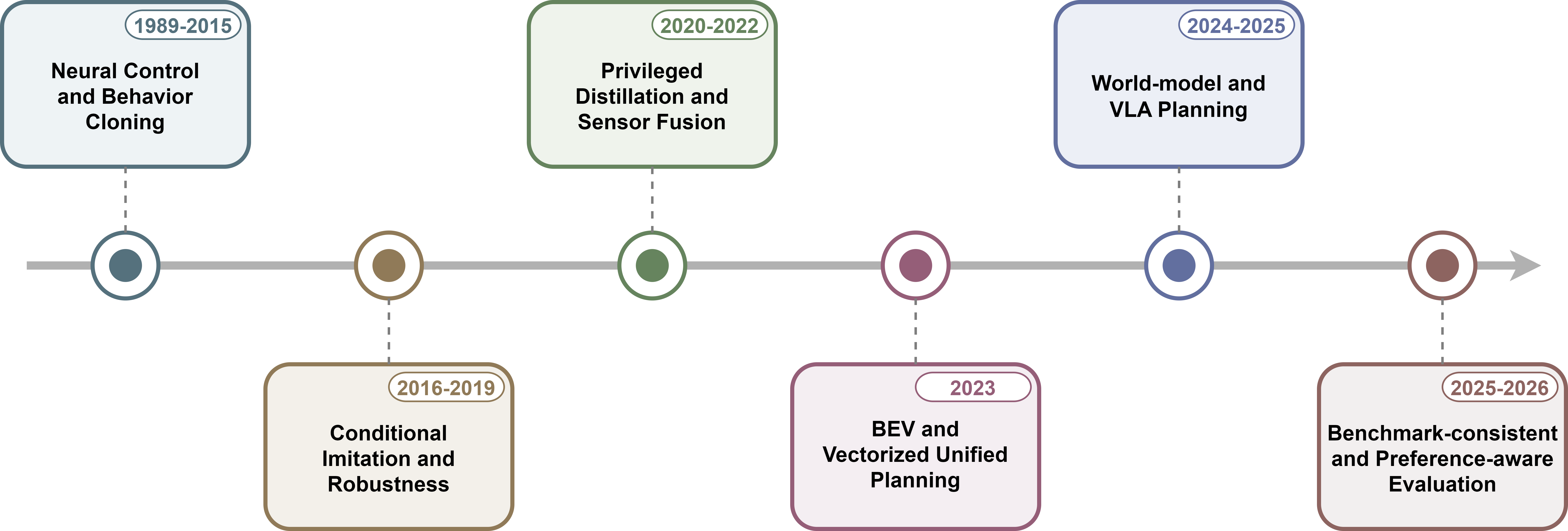}
\caption{Conceptual roadmap of planning-oriented E2E-AD. The field has evolved from direct control imitation toward structured planning interfaces, consequence-aware models, action-grounded language systems, and evaluation protocols that assess planning quality beyond trajectory similarity.}
\label{fig:historical_roadmap}
\end{figure*}

Figure~\ref{fig:historical_roadmap} summarizes the main historical trend reviewed in this section. The evolution of E2E-AD is not a single architectural lineage, but a shift from imitating low-level control toward learning representations, supervision signals, and evaluation protocols that make planning behavior more inspectable and benchmarkable.

\subsection{From Neural Control to Conditional Imitation}
The roots of E2E driving predate modern deep learning. ALVINN showed that a neural network could map road images to steering commands \cite{pomerleau1988alvinn}, and later off-road driving systems demonstrated direct perception-action learning under constrained conditions \cite{muller2005off}. DeepDriving introduced affordance prediction as an intermediate representation between hand-engineered perception and direct control \cite{chen2015deepdriving}. PilotNet then became the modern reference point for camera-to-steering learning by showing that a convolutional network trained on human demonstrations could control a vehicle in real-world road scenes \cite{bojarski2016end}.

A key limitation of direct behavior cloning (BC) is controllability. The same observation can require different actions depending on navigation intent, such as going straight, turning left, or turning right. Conditional imitation learning addressed this ambiguity by conditioning the policy on high-level commands \cite{codevilla2018end}. This made E2E policies more suitable for urban driving, but did not remove distribution shift. Once a learned policy deviates from expert behavior, it may enter states that are rare in the training data, causing errors to compound into off-route behavior or collisions. The NoCrash benchmark and related BC analyses made this failure mode explicit \cite{codevilla2019exploring}. ChauffeurNet responded with synthesized perturbations and auxiliary losses, showing that robustness depends not only on network design but also on how recovery states are represented during training \cite{bansal2018chauffeurnet,chen2019deep}.

\subsection{Privileged Distillation and Robust Imitation}
A second phase used privileged information to improve imitation learning. Learning by Cheating trained a privileged teacher with access to ground-truth state and distilled it into an image-based student \cite{chen2020learning}. This approach was influential because it preserved the deployment interface of an E2E sensor policy while exploiting structured supervision during training. Other studies expanded the source of imitation signals beyond ego demonstrations. Learning by Watching extracted supervision from observed non-ego agents \cite{zhang2021learning}, while Learning From All Vehicles used trajectories of all observed vehicles rather than only the data-collection ego vehicle \cite{chen2022learning}. These methods highlight a recurring lesson: E2E driving performance depends as much on supervision design as on network architecture.

Reinforcement learning (RL) has also contributed to E2E driving, although it has been less dominant than imitation learning \cite{kiran2021deep,aradi2020survey,wu2024recent}. CIRL combined imitation and RL for controllable visual driving \cite{liang2018cirl}. Model-free RL with implicit affordances showed that RL can operate in urban scenarios when the policy receives planning-relevant structure \cite{toromanoff2020end,chen2021interpretable}. Urban Driver explored offline policy gradients using a differentiable simulator built from real-world demonstrations and maps \cite{scheel2022urban}. These studies suggest that RL is most useful not as a replacement for imitation, but as a way to address closed-loop interaction, reward design, and recovery from policy-induced errors \cite{chen2019model,zhu2020safe}.

\subsection{From Sensor Fusion to Unified Planning}
The next phase shifted E2E-AD from direct control prediction toward richer planning outputs \cite{wang2021learning}. TransFuser demonstrated the value of multimodal transformer fusion for E2E driving \cite{prakash2021multi}, while NEAT used neural attention fields to impose more explicit scene structure \cite{chitta2021neat}. TCP linked trajectory prediction with low-level control and became a strong CARLA baseline \cite{wu2022trajectory}. PlanT used object-level representations for explainable planning \cite{renz2023plant}, MILE formulated urban driving through model-based imitation learning \cite{hu2022model}, and InterFuser incorporated interpretable sensor fusion with safety-oriented constraints \cite{shao2023safety}.

UniAD marked a conceptual turning point by organizing detection, tracking, mapping, prediction, occupancy, and planning around the final planning objective \cite{hu2023planning}. Rather than treating upstream tasks as independent goals, UniAD argued that perception and prediction should be optimized to support planning. VAD and VADv2 further advanced this direction through vectorized scene representations and probabilistic planning \cite{jiang2023vad,jiang2024vadv2}. DriveAdapter examined the coupling barrier between perception and planning \cite{jia2023driveadapter}, ThinkTwice showed that planning decoder capacity can limit driving performance \cite{jia2023think}, and PARA-Drive targeted real-time parallelized deployment \cite{weng2024drive}. By this stage, the central question had shifted from whether a network can imitate steering to which representation, decoder, supervision signal, and evaluation protocol can produce safe and reliable plans.

\begin{figure*}[t]
\centering
\includegraphics[width=0.90\textwidth]{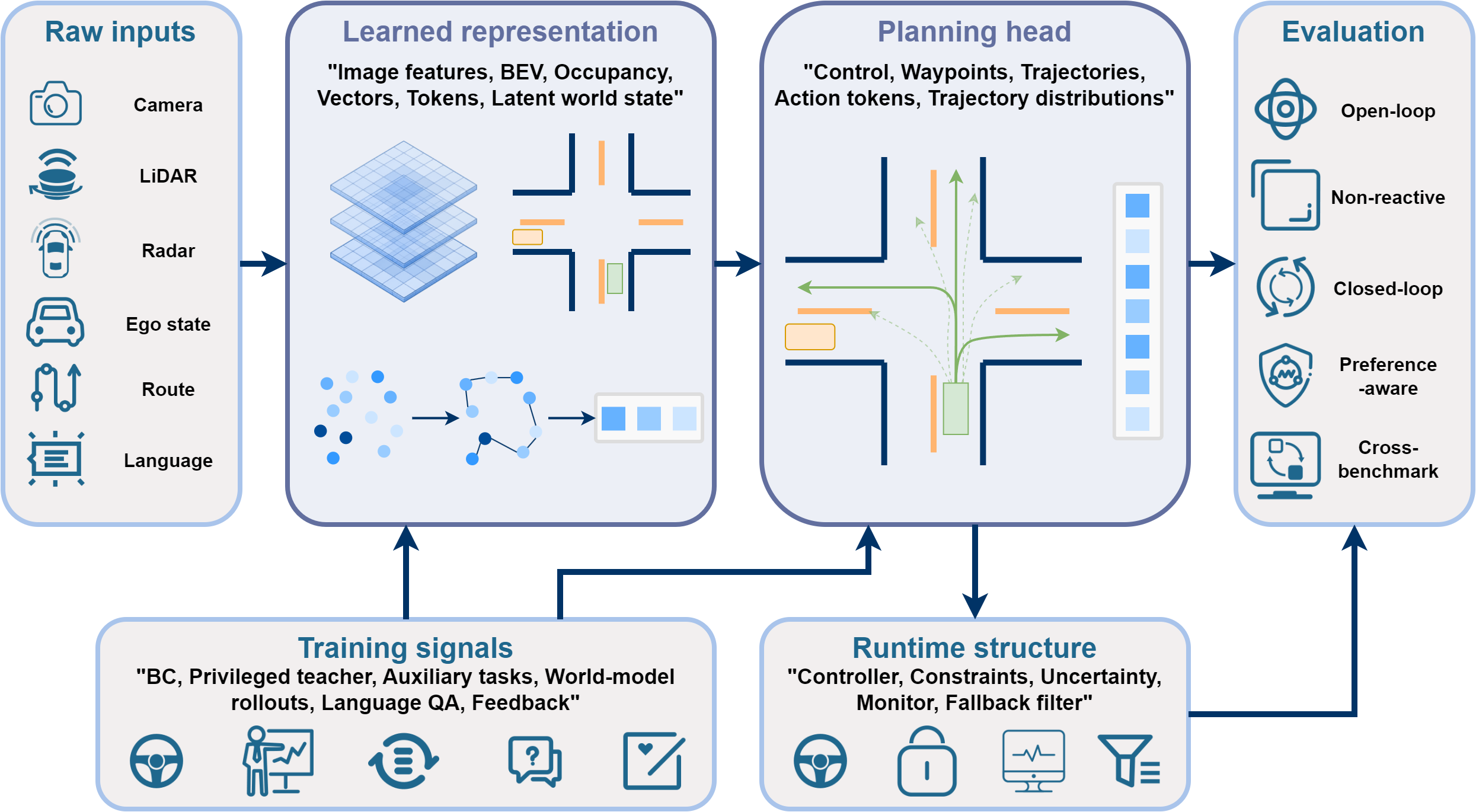}
\caption{A planning-oriented view of end-to-end autonomous driving. The key distinction is whether the structure is learned and optimized for downstream planning and evaluated under protocols that reflect driving quality.}
\label{fig:taxonomy}
\end{figure*}

\section{Planning-Oriented Taxonomy}

Figure~\ref{fig:taxonomy} summarizes the taxonomy used in this survey. We categorize planning-oriented E2E-AD methods along four axes: input representation, output space, supervision signal, and evaluation protocol. Table~\ref{tab:axes} makes these axes explicit. This taxonomy avoids a common limitation of classifying E2E-AD methods only by backbone names. A transformer encoder, diffusion decoder, or VLM is not by itself a driving formulation. What matters is which driving variables are exposed to the policy, which output space the policy optimizes, what supervision makes the output learnable, and which evaluation protocol supports the reported claim. Two methods with different backbones may belong to the same planning family if they predict similar trajectories under similar supervision. Conversely, two transformer-based systems may address different problems if one is a BEV planner trained by privileged distillation and the other is a VLA planner trained by language-action alignment.

\begin{table*}[t]
\caption{Four-Axis Taxonomy of Planning-Oriented E2E-AD.}
\label{tab:axes}
\centering
\setlength{\extrarowheight}{2pt}
\begin{tabularx}{\textwidth}{p{0.15\textwidth}Y Y Y}
\toprule
Axis & Main options & Representative works & Planning implication \\
\midrule
Input representation & Front-view image; multi-view image; LiDAR/radar fusion; BEV or occupancy; object/vector tokens; latent world state; VLM/VLA token space & PilotNet \cite{bojarski2016end}, TransFuser \cite{prakash2021multi}, UniAD \cite{hu2023planning}, VAD \cite{jiang2023vad}, LAW \cite{li2025enhancing}, SpaceDrive \cite{li2026spacedrive} & Determines geometry, interaction modeling, map usage, and whether the policy can reason about unseen or rare agents. \\
Output space & Steering/throttle; waypoints; trajectory; multi-modal trajectory distribution; scene rollout; language-conditioned action tokens & CIL \cite{codevilla2018end}, TCP \cite{wu2022trajectory}, VADv2 \cite{jiang2024vadv2}, DiffusionDrive \cite{liao2025diffusiondrive}, WAM-Flow \cite{xu2026wam} & Determines controllability, uncertainty representation, controller dependence, and metric compatibility. \\
Supervision & Human BC; synthesized perturbations; privileged teacher; rule-based or RL expert; auxiliary perception/prediction; world-model prediction; language QA; rater preference & ChauffeurNet \cite{bansal2018chauffeurnet}, Learning by Cheating \cite{chen2020learning}, Hydra-MDP \cite{li2024hydra}, LEAD \cite{nguyen2026lead}, DriveLM \cite{sima2024drivelm}, WOD-E2E \cite{xu2026wod} & Determines whether the policy learns recovery, safety margins, semantic intent, and long-tail preferences. \\
Evaluation protocol & Open-loop log replay; non-reactive simulation; reactive closed-loop simulation; real-world closed-loop testing; long-tail and preference-aware evaluation; cross-benchmark correlation & nuScenes \cite{caesar2020nuscenes}, nuPlan \cite{caesar2021nuplan}, NAVSIM \cite{dauner2024navsim}, Bench2Drive \cite{jia2024bench2drive}, Fail2Drive \cite{gerstenecker2026fail2drive}, WOD-E2E \cite{xu2026wod} & Determines whether claims concern imitation fidelity, scalable proxy planning, interaction, generalization, or human-aligned judgment. \\
\bottomrule
\end{tabularx}
\end{table*}

\subsection{Input Representation}
Early E2E systems relied mainly on front-view images \cite{bojarski2016end,codevilla2018end,zhao2022attention}. Modern methods increasingly use multi-camera inputs, BEV features, occupancy grids, object queries, vectorized map and agent tokens, latent world states, or VLM/VLA token spaces \cite{prakash2021multi,hu2023planning,jiang2023vad,li2025end,chen2020end}. BEV and vectorized representations are attractive because they encode geometry, lanes, maps, and interactions in a coordinate frame closer to motion planning \cite{li2024context,wang2025nest}. Sparse token designs, such as PlanT and DriveTransformer, further reduce the cost of dense spatial processing while retaining object-level structure \cite{renz2023plant,jia2025drivetransformer,liao2024mftraj}. For VLM-based systems, the central question is whether language and spatial tokens improve action generation, rather than only producing plausible explanations \cite{sima2024drivelm,li2026spacedrive,zhou2026opendrivevla,liao2024gpt}.

\subsection{Output Space}
The output space determines how a learned policy connects perception to executable driving behavior. Direct steering and throttle commands minimize post-processing, but they are difficult to constrain and interpret. Waypoints and trajectories provide a more inspectable planning interface because they can be tracked by a controller and checked against drivable-area, collision, comfort, and route constraints \cite{gupta2023interaction,zhou2019state,liao2025sa}. More recent systems predict trajectory distributions, scene rollouts, or action tokens to represent multiple valid futures in interactive scenes \cite{feng2025artemis}. VADv2, DiffusionDrive, World4Drive, and WAM-Flow reflect this trend toward probabilistic or generative planning \cite{jiang2024vadv2,liao2025diffusiondrive,zheng2025world4drive,xu2026wam}.

\subsection{Supervision Signal}
Although imitation learning from human or expert trajectories remains the dominant supervision source, planning-oriented E2E-AD has diversified its training signals. Perturbation training and data aggregation expose policies to recovery states \cite{bansal2018chauffeurnet,prakash2020exploring}; privileged distillation uses ground-truth or simulator-state teachers \cite{chen2020learning}; and auxiliary perception, prediction, and occupancy losses stabilize planning-oriented representations \cite{hu2023planning,shao2023safety}. World-model objectives add future prediction as a self-supervised signal \cite{min2024driveworld,li2025enhancing,li2025end}, while language-based systems introduce question answering, instruction following, commentary, or action-language alignment \cite{sima2024drivelm,renz2025simlingo,fu2025orion,li2025generative}. Recent work such as LEAD further shows that expert design should account for what the student policy can observe and imitate \cite{nguyen2026lead,liao2025cot}.

\subsection{Evaluation Protocol}
Evaluation is the axis on which many reported advances become difficult to compare. Open-loop trajectory matching measures similarity to logged human behavior. It is scalable and reproducible, but it cannot measure recovery from the policy's own mistakes and may reward ego-state shortcuts \cite{li2024ego}. Closed-loop simulation captures interaction and recovery but can suffer from simulator domain gap, stochasticity, and high computational cost. Non-reactive real-log simulation, as in NAVSIM, trades full interaction for scale and real sensor realism \cite{dauner2024navsim}. Long-tail and preference-aware benchmarks, such as WOD-E2E, add rare-scenario focus and rater feedback labels \cite{xu2026wod}. A planning-oriented survey should therefore compare results only within aligned protocols and should separate open-loop, non-reactive, reactive closed-loop, and preference-aware claims.

We summarize representative planning-oriented E2E-AD methods in Table~\ref{tab:method_level}. Code and benchmark status are recorded according to publicly available information as of June 2026. The term ``limited'' denotes evaluation evidence that is mainly open-loop, non-reactive, or benchmark-constrained, rather than full reactive closed-loop driving. It should not be interpreted as a judgment of real-world deployability. The public-resource labels are defined as follows: \emph{Code} denotes a runnable public implementation, \emph{Partial} denotes an incomplete or version-sensitive release, \emph{Benchmark} denotes public evaluation resources or benchmark baselines, and \emph{Paper-only} denotes the absence of comparable public resources.

\begin{table*}[t]
\caption{Representative Planning-Oriented E2E-AD Methods.}
\label{tab:method_level}
\centering
\setlength{\extrarowheight}{2pt}
\scriptsize
\begin{tabularx}{\textwidth}{p{0.15\textwidth}p{0.045\textwidth}Y p{0.16\textwidth}Y p{0.08\textwidth}}
\toprule
Method & Year & Planning Interface & Training Signal & Evaluation Evidence & Public Resource \\
\midrule
PilotNet \cite{bojarski2016end} & 2016 & Front camera $\rightarrow$ Image features $\rightarrow$ Steering & BC from demonstrations & Real-road closed-loop demonstration & Partial \\
CIL \cite{codevilla2018end} & 2018 & Camera + command $\rightarrow$ Image features $\rightarrow$ Control & Command-conditioned BC & CARLA closed-loop routes & Code \\
ChauffeurNet \cite{bansal2018chauffeurnet} & 2019 & Rasterized scene $\rightarrow$ Context features $\rightarrow$ Trajectory & BC with perturbation and auxiliary losses & Internal/RSS-style limited evidence & Paper-only \\
Learning by Cheating \cite{chen2020learning} & 2020 & Camera student $\rightarrow$ Distilled waypoints/control & Privileged teacher-student IL & CARLA closed-loop routes & Code \\
TransFuser \cite{prakash2021multi} & 2021 & Camera + LiDAR $\rightarrow$ Fusion transformer $\rightarrow$ Waypoints/control & BC with sensor fusion & CARLA closed-loop routes & Code \\
TCP \cite{wu2022trajectory} & 2022 & Camera $\rightarrow$ Shared features $\rightarrow$ Trajectory and control & BC with coupled planning/control & CARLA closed-loop routes & Code \\
PlanT \cite{renz2023plant} & 2022 & Objects + route $\rightarrow$ Object tokens $\rightarrow$ Plan/control & BC over object-level scenes & CARLA closed-loop routes & Code \\
UniAD \cite{hu2023planning} & 2023 & Multi-view cameras $\rightarrow$ BEV/query stack $\rightarrow$ Trajectory & Multi-task planning supervision & nuScenes open-loop/limited planning evidence & Code \\
VAD \cite{jiang2023vad} & 2023 & Multi-view cameras $\rightarrow$ Vectorized scene $\rightarrow$ Trajectory & Multi-task vectorized planning & nuScenes open-loop/limited planning evidence & Code \\
DriveAdapter \cite{jia2023driveadapter} & 2023 & Multi-view cameras $\rightarrow$ Adapted BEV/planning features $\rightarrow$ Trajectory & Modular-to-E2E adaptation & nuScenes open-loop/limited planning evidence & Code \\
PARA-Drive \cite{weng2024drive} & 2024 & Multi-sensor inputs $\rightarrow$ Parallel unified stack $\rightarrow$ Planning outputs & Multi-task unified learning & nuScenes limited planning evidence & Partial \\
NAVSIM agents \cite{dauner2024navsim} & 2024 & Real sensor logs $\rightarrow$ Varied encoders $\rightarrow$ Trajectory & Varied training recipes & NAVSIM non-reactive real-log evaluation & Benchmark \\
Bench2Drive agents \cite{jia2024bench2drive} & 2024 & CARLA sensors $\rightarrow$ Varied agents $\rightarrow$ Control/trajectory & IL/RL and expert data & Bench2Drive closed-loop benchmark & Benchmark \\
VADv2 \cite{jiang2024vadv2} & 2024 & Multi-view cameras $\rightarrow$ Vectorized scene $\rightarrow$ Probabilistic plan & Multi-task probabilistic planning & nuScenes open-loop/limited planning evidence & Partial \\
EMMA \cite{hwang2024emma} & 2025 & Camera + route/text prompts $\rightarrow$ Multimodal tokens $\rightarrow$ Text-form plan & Multitask prompting & Waymo/internal offline evidence & Paper-only \\
DriveTransformer \cite{jia2025drivetransformer} & 2025 & Multi-view cameras $\rightarrow$ Sparse task queries $\rightarrow$ Trajectory & Unified transformer supervision & nuScenes open-loop/limited planning evidence & Code \\
SimLingo \cite{renz2025simlingo} & 2025 & Vision + language $\rightarrow$ VLA tokens $\rightarrow$ Action/trajectory & Language-action alignment & CARLA closed-loop routes & Code \\
WoTE \cite{li2025end} & 2025 & Multi-view cameras $\rightarrow$ BEV world model $\rightarrow$ Scored trajectory & World-model evaluator & nuScenes/CARLA limited evidence & Partial \\
LEAD \cite{nguyen2026lead} & 2026 & Camera + route $\rightarrow$ Student-aware features $\rightarrow$ Control/trajectory & Expert-aligned IL & CARLA closed-loop routes & Code \\
WOD-E2E \cite{xu2026wod} & 2026 & 8-camera logs $\rightarrow$ Varied planners $\rightarrow$ Trajectory & Varied methods with preference-aware evaluation & WOD-E2E long-tail open-loop/preference & Benchmark \\
\bottomrule
\end{tabularx}
\end{table*}

Across the main method families, each generation solved one bottleneck while revealing another. As summarized in Table~\ref{tab:tradeoffs}, direct BC made learning feasible but exposed covariate shift. Conditional imitation improved controllability but did not solve recovery. Privileged distillation improved supervision but introduced learner-expert asymmetry. BEV and vector methods improved geometric reasoning but increased dependence on intermediate labels, maps, and expensive encoders. World models brought consequence modeling but raised calibration and rollout-validity questions \cite{gao2024enhance,guan2025world}. VLA systems added semantic reasoning but created a new alignment problem between text-space reasoning and metric-space control \cite{zhou2026opendrivevla}.

\begin{table*}[t]
\caption{Design Trade-Offs in Planning-Oriented E2E-AD.}
\label{tab:tradeoffs}
\centering
\setlength{\extrarowheight}{2pt}
\begin{tabularx}{\textwidth}{p{0.19\textwidth}Y Y Y}
\toprule
Design choice & What it improves & What it risks & Useful diagnostic \\
\midrule
Direct control output & Minimal engineering and fast inference. & Weak interpretability, unstable actuation, hard safety checks, poor multi-modality. & Compare closed-loop recovery and comfort against waypoint or trajectory variants. \\
Waypoint or trajectory output & Better compatibility with controllers, constraints, and planning metrics. & Controller choices can hide policy weaknesses; open-loop trajectory loss can over-penalize valid alternatives. & Report controller details and evaluate both trajectory error and closed-loop infractions. \\
Dense BEV/occupancy representation & Strong geometry, map alignment, and interaction modeling. & High compute, sensitivity to perception labels, possible overfitting to static map priors. & Ablate maps, history length, and sensor modality under closed-loop routes. \\
Sparse object/vector tokens & Efficient interaction reasoning and clearer object-level structure. & May miss unmodeled context, road texture, small obstacles, or non-canonical objects. & Stress test unusual agents, free-space reasoning, and mapless scenarios. \\
Privileged teacher & Strong supervision and scalable simulated data. & Teacher may rely on information unavailable to the student. & Measure learner-expert asymmetry and student performance under partial observability. \\
Generative or probabilistic planner & Represents multiple valid futures and avoids mode averaging. & Hard to rank samples; can generate plausible but unsafe trajectories. & Report sample diversity, best-of-N versus selected trajectory, and safety-progress trade-offs. \\
World-model rollout & Enables consequence-aware planning and self-supervised future learning. & Latent rollouts can be miscalibrated in rare cases. & Evaluate world-model prediction quality in long-tail scenes and compare selected plans with closed-loop outcomes. \\
Language/VLA reasoning & Adds semantic context, instruction following, and explainability. & Explanation-action mismatch, latency, hallucination, and weak metric grounding. & Separate language-at-training, language-at-inference, and language-as-explanation ablations. \\
\bottomrule
\end{tabularx}
\end{table*}

\section{Architectures and Learning Paradigms}
\subsection{Control, Waypoints, and Trajectories}
Direct control prediction is attractive because it minimizes the number of engineering components between perception and actuation. Yet it gives the network little room to express uncertainty or to separate strategic planning from actuator-level control. Waypoint and trajectory prediction provide a more useful intermediate action space \cite{zhang2023predictive}. A waypoint sequence can be supervised from logs, converted to control by a controller, and checked for collisions, drivable-area violations, or comfort \cite{rasekhipour2016potential}. TCP exploited this by jointly predicting trajectories and controls, becoming a strong CARLA baseline \cite{wu2022trajectory}. ThinkTwice showed that the planning decoder itself can be a bottleneck and that look-ahead refinement improves driving quality \cite{jia2023think}.

The choice of output also shapes evaluation. A trajectory can be scored by displacement error, collision proxies, route progress, time-to-collision, comfort, or preference labels. A direct control command is harder to compare under open-loop logs because the same control may have different consequences depending on the vehicle state and controller dynamics. As benchmarks have become more planning-centric, trajectory outputs have become the dominant interface for E2E research.

There is also an underappreciated mismatch between the supervised target and the deployed behavior. A logged trajectory is not a universal optimum; it is one human driver's realization under one interaction history. If the model deviates slightly, the logged future may no longer be feasible or even desirable. This is why pure pointwise L2 trajectory loss can be misleading. It rewards imitation of a single future but does not teach the policy how to choose among safe alternatives after perturbation. Modern planners increasingly add multi-modal heads, trajectory refinement, rule-aware costs, or learned scoring functions to bridge this gap \cite{li2024hydra,li2025hydra,liao2025diffusiondrive,xu2026wam,yao2026drivesuprim}.

Overall, the shift from direct control to trajectory output is not merely a change in output format; it makes the policy easier to inspect through controllers, safety constraints, and planning metrics, while exposing new questions about multi-modal futures and plan selection.

\subsection{BEV, Occupancy, and Vectorized Scene Representations}
BEV representations became popular because they provide a geometry-aware coordinate frame for planning. They can integrate multi-view cameras, maps, lanes, objects, and occupancy into a common spatial grid. ST-P3, UniAD, and related works use BEV-like structures to support detection, prediction, occupancy, and planning losses \cite{hu2022st,hu2023planning}. Vectorized representations go further by representing lanes, agents, and trajectories as structured tokens rather than dense grids. VAD demonstrated that vectorized scene representation can support efficient autonomous driving \cite{jiang2023vad}; VADv2 extended this direction with probabilistic planning \cite{jiang2024vadv2}.

The planning advantage of BEV and vector spaces is interpretability and constraint compatibility. A planner can reason about drivable area, lane topology, agent interaction, and route progress more naturally in a spatial representation than in raw image features. The risk is that dense BEV processing can be expensive and may inherit the limitations of the perception labels used to supervise it. Recent sparse and register-based designs, including DriveTransformer and DrivoR, can be understood as attempts to retain planning-relevant structure while reducing computational cost \cite{jia2025drivetransformer,kirby2026driving}.

Overall, the value of BEV, occupancy, and vector representations is not only stronger perception; their main planning value is to convert raw sensing into geometry that can be checked against maps, agents, free space, and route constraints.

\subsection{Unified Stacks and Task Coupling}
UniAD's importance lies less in a single module than in its planning-oriented philosophy: upstream tasks should be prioritized and designed according to their contribution to planning \cite{hu2023planning}. This stands in contrast to modular pipelines where detection, tracking, mapping, and prediction are optimized separately before being handed to a planner. Unified architectures can reduce interface mismatch and allow gradient signals from planning to shape representation learning \cite{zhang2026perception}. DriveAdapter highlighted the coupling barrier that arises when perception and planning are trained or adapted separately \cite{jia2023driveadapter}. DriveTransformer further simplified the architecture by using task self-attention, sensor cross-attention, and temporal cross-attention as unified operations \cite{jia2025drivetransformer}.

Task coupling also creates new risks. If a unified model performs well, it may be unclear which internal component is responsible. If it fails, debugging can be harder than in a modular stack. Moreover, joint training can allow shortcut learning: a model may rely excessively on ego state or route signals while underusing visual evidence \cite{li2024ego}. Planning-oriented architecture must therefore be paired with causal diagnostics, ablations, and benchmark protocols that expose shortcut behavior.

A recurring architectural question is therefore not simply whether to be modular or end-to-end. It is where to place the interfaces that must remain stable for safety, debugging, and evaluation. A system can be end-to-end trained while retaining explicit interfaces such as occupancy, lane topology, object tokens, route commands, uncertainty estimates, or candidate trajectories. Conversely, a system can appear modular but still propagate downstream losses into upstream representations. Across recent systems, a common pattern is to expose variables needed for planning and safety checks while reducing hand-designed objectives that dominate the learned policy.

\subsection{Uncertainty and Generative Planning}
Driving decisions are not single-label predictions \cite{liao2024cognitive}. At an unprotected turn, a vehicle may wait, creep, or proceed depending on subtle social and dynamic cues \cite{liao2024bat,liao2024human}. Deterministic regression to the logged trajectory can punish valid alternatives and average over modes. Probabilistic and generative planners address this by representing multiple candidate futures \cite{yin2026diffrefiner}. VADv2 models planning as a probabilistic distribution over actions \cite{jiang2024vadv2}. DiffusionDrive brings diffusion-style generation to trajectory planning \cite{liao2025diffusiondrive}. Momentum-aware planning stabilizes trajectory generation by incorporating perception and trajectory momentum \cite{song2025don}. WAM-Flow casts trajectory planning as discrete flow matching, offering a parallel coarse-to-fine alternative to autoregressive decoding \cite{xu2026wam}.

The open question is how to evaluate generative plans. A distribution is useful only if it ranks safe and goal-consistent trajectories above unsafe or irrelevant ones. World-model-based evaluation, preference labels, and closed-loop simulation are therefore natural complements to generative planning.

\subsection{Architecture Lessons for Closed-Loop Behavior}
Closed-loop performance depends on design details that may be invisible in an open-loop table. First, policies need fast local reactions and longer-horizon route consistency at the same time. A model that optimizes long-horizon L2 may respond too slowly to sudden cut-ins; a model that over-optimizes short-term collision avoidance may freeze or crawl. Hydra-NeXt makes this tension explicit by combining trajectory prediction, control prediction, and trajectory refinement \cite{li2025hydra}. Second, history aggregation must be aligned with planning horizons. BridgeAD argues that historical prediction and future planning should be connected at the query level rather than treated as generic temporal fusion \cite{zhang2025bridging}. Third, closed-loop performance is sensitive to whether the policy has learned free space, occupied space, and route intent as concepts rather than as dataset correlations. Fail2Drive reports that strong models can degrade sharply under paired distribution shifts, revealing failures that aggregate driving score hides \cite{gerstenecker2026fail2drive}.

These observations suggest a practical checklist for architecture papers. Authors should report whether the planner uses explicit route tokens or target points, whether it predicts speed in addition to lateral path, whether the controller is learned or fixed, whether trajectory samples are ranked by a learned cost, and whether the policy can recover when the ego state deviates from the logged trajectory \cite{du2022comfortable}. Without these details, architecture comparisons often become comparisons of hidden evaluation machinery.

Overall, architectural novelty is difficult to interpret without reporting the planning interface, controller, safety wrapper, and recovery mechanism. These details often explain closed-loop behavior as much as the encoder backbone.

\subsection{Training Signals and Data Engines}
\subsubsection{Behavior Cloning Is Necessary but Insufficient}
As shown in Table \ref{tab:training}, BC remains the most scalable supervision source because human driving logs are abundant relative to interactive safety-critical trials. However, the literature repeatedly shows that vanilla BC is fragile under covariate shift \cite{codevilla2019exploring}. The key challenge is that expert data is collected under expert state distributions, while the learner induces its own state distribution at test time. Perturbation training, synthesized failures, data aggregation, and feedback-guided correction attempt to close this gap \cite{bansal2018chauffeurnet,prakash2020exploring,zhang2024feedback}.

The learner-expert mismatch becomes especially important in simulation. Privileged experts can generate large datasets cheaply, but an expert that uses unobservable ground-truth state may produce demonstrations that a camera-based student cannot reliably imitate. LEAD reframes this issue by designing the expert and navigation conditioning to reduce visibility, uncertainty, and intent asymmetry between teacher and student \cite{nguyen2026lead}. This is a useful conceptual advance because it treats expert design as part of the learning problem rather than as a fixed data source.

A recurring lesson is that demonstration quality is not the same as expert driving quality. A perfect expert under full observability can be a poor teacher for a limited sensor policy if it makes decisions using information that the student cannot infer. Conversely, a slightly conservative expert with realistic sensing assumptions may produce more learnable supervision. This matters for CARLA-style data generation, where the community often trains on rule-based or RL-based expert rollouts. This survey therefore distinguishes between \emph{expert optimality}, \emph{expert learnability}, and \emph{student deployability}. LEAD makes this distinction explicit; Hydra-MDP is complementary because it uses multiple teachers and multi-target distillation to expose diverse planning objectives \cite{li2024hydra}.

\subsubsection{Auxiliary Tasks and Privileged Distillation}
Auxiliary tasks are often criticized as making E2E systems less ``pure.'' In practice, they are one of the main reasons modern systems work. Detection, segmentation, occupancy, motion forecasting, route prediction, and language QA can regularize representation learning and provide denser gradients than final trajectory loss alone. Learning by Cheating formalized one version of this idea through privileged distillation \cite{chen2020learning}. UniAD, InterFuser, and VAD use structured intermediate supervision to support planning \cite{hu2023planning,shao2023safety,jiang2023vad}. The planning-oriented stance is that auxiliary tasks are justified when they improve the final plan and are evaluated under planning metrics.

A useful distinction is between \emph{auxiliary supervision} and \emph{auxiliary interfaces}. An auxiliary loss may be used only during training, after which the deployed system outputs a trajectory. An auxiliary interface remains visible at inference time, such as occupancy, object tokens, map elements, or language rationales. Training-only auxiliary tasks can improve representation learning without increasing deployment complexity. Inference-time interfaces can improve debugging and safety monitoring but may increase latency and require additional calibration. Papers should report which auxiliary signals are removed at inference and which remain part of the deployed policy.

\subsubsection{World-Model Supervision}
World models introduce a different supervision source: predict the future state of the driving scene and use that predictive model to support planning. MILE explored model-based imitation for urban driving \cite{hu2022model}. DriveWorld used world-model pretraining for 4D scene understanding \cite{min2024driveworld}. LAW used a latent world model to enhance E2E driving \cite{li2025enhancing}. WoTE uses a BEV world model to evaluate candidate trajectories online \cite{li2025end}, while World4Drive constructs intention-aware physical latent representations for planning without manual perception annotations \cite{zheng2025world4drive}. These methods shift the training question from ``what action did the expert take?'' to ``what consequences follow from candidate actions?''

This shift is important but not yet settled. Latent futures may improve planning, but they can also become inaccurate precisely in rare situations where planning most needs reliability \cite{kalra2016driving,liao2024physics}. Future work should report not only trajectory scores but also world-model calibration, uncertainty, failure cases, and sensitivity to distribution shift.

\subsubsection{Language as Supervision, Interface, and Planner}
Language enters E2E driving in at least three ways. First, it can provide structured supervision, as in DriveLM's graph visual question answering \cite{sima2024drivelm}. Second, it can serve as an interface for high-level instruction or explanation, as in LMDrive, DriveVLM, and VLP \cite{shao2024lmdrive,tian2025drivevlm,pan2024vlp}. Third, it can become part of the action-generation mechanism, as in SimLingo, ORION, SpaceDrive, and WAM-Flow \cite{renz2025simlingo,fu2025orion,li2026spacedrive,xu2026wam}.

The key evaluation question is whether language improves driving behavior. A model that produces persuasive explanations but weak trajectories is not a better driving policy. SimLingo is important because it explicitly aligns language understanding with action behavior \cite{renz2025simlingo}. DiMA is important from a systems perspective because it distills multimodal LLM knowledge into a vision planner, making the language model optional at inference time \cite{hegde2025distilling}. These works suggest that language can be useful as a training-time teacher, an inference-time reasoner, or a post-hoc explanation mechanism, but these roles should not be conflated.

\subsubsection{Post-Training, Feedback, and Preference Signals}
The most recent frontier borrows ideas from foundation-model post-training. Feedback-guided AD uses language feedback to explain prediction failures and refine behavior cloning \cite{zhang2024feedback}. WOD-E2E introduces rater preference labels for long-tail scenes, shifting evaluation away from pure distance matching \cite{xu2026wod}. WAM-Flow includes simulator-guided alignment through rewards for safety, progress, and comfort \cite{xu2026wam}. These works suggest that future E2E-AD training may resemble a multi-stage process: pretrain representations on large driving video or multimodal corpora, train a planner by imitation, post-train with safety and preference feedback, and finally validate under closed-loop and long-tail protocols.

This direction is powerful but delicate. Preference labels can encode human judgment about rare interactions, but they also depend on rater instructions and scenario framing. Reward-based post-training can improve benchmark scores, but it may overfit to metric thresholds. Language feedback can expose causal mistakes, but it can also inject noisy rationales. A high-quality survey should therefore treat feedback and preference learning as an emerging research program, not as a solved replacement for closed-loop validation \cite{liao2025chain}.

\begin{table*}[t]
\caption{Training Signals for Planning-Oriented E2E-AD.}
\label{tab:training}
\centering
\setlength{\extrarowheight}{2pt}
\begin{tabularx}{\textwidth}{p{0.19\textwidth}Y Y Y}
\toprule
Training signal & Representative works & Benefit & Main risk \\
\midrule
Human behavior cloning & PilotNet \cite{bojarski2016end}, CIL \cite{codevilla2018end} & Simple, scalable, log-based supervision. & Covariate shift, route ambiguity, multimodal futures collapsed into one label. \\
Perturbation and data aggregation & ChauffeurNet \cite{bansal2018chauffeurnet}, data aggregation \cite{prakash2020exploring}, feedback-guided AD \cite{zhang2024feedback} & Exposes policy to recovery states and corrective examples. & Quality depends on perturbation design and feedback reliability. \\
Privileged teacher-student learning & Learning by Cheating \cite{chen2020learning}, LEAD \cite{nguyen2026lead} & Uses simulator or privileged state to produce strong supervision. & Expert may rely on information unavailable to the deployed student. \\
Auxiliary perception/prediction tasks & UniAD \cite{hu2023planning}, InterFuser \cite{shao2023safety}, VAD \cite{jiang2023vad} & Stabilizes learning and improves interpretability. & May overfit to intermediate metrics rather than planning quality. \\
World-model prediction & DriveWorld \cite{min2024driveworld}, LAW \cite{li2025enhancing}, WoTE \cite{li2025end}, World4Drive \cite{zheng2025world4drive} & Learns consequences and supports candidate trajectory evaluation. & Latent futures may be poorly calibrated in long-tail scenes. \\
Language and preference supervision & DriveLM \cite{sima2024drivelm}, SimLingo \cite{renz2025simlingo}, WOD-E2E \cite{xu2026wod} & Adds reasoning labels, action alignment, and human preference signals. & Risk of explanation-action mismatch and annotation subjectivity. \\
\bottomrule
\end{tabularx}
\end{table*}

% \FloatBarrier
\section{World Models and VLA Driving}
\subsection{Why World Models Matter for Planning}
Planning is about consequences. A purely reactive policy can imitate expert actions, but it does not explicitly ask what will happen if the ego vehicle accelerates, yields, nudges forward, or changes lane. World models attempt to fill this gap by learning latent dynamics of the scene.  As detailed in Table~\ref{tab:worldmodels}, GenAD framed autonomous driving as generalized predictive modeling and introduced OpenDV-2K as a driving video resource \cite{yang2024generalized}. Driving into the Future connected multi-view visual forecasting with planning \cite{wang2024driving}. LAW and WoTE show that latent or BEV futures can serve as planning supervision or trajectory evaluation mechanisms \cite{li2025enhancing,li2025end}. World4Drive pushes the idea further by using intention-aware physical latent world states to generate and rank trajectories \cite{zheng2025world4drive}.

The planning benefit is intuitive: a policy should prefer trajectories whose predicted consequences are safe, smooth, and goal-consistent. The unresolved issue is validation. A world model can appear visually or metrically plausible while missing rare interactions, unusual agents, or causal effects of ego behavior. This is why world-model papers should be evaluated not only on reconstruction or prediction losses, but also on closed-loop planning outcomes, rare-scenario behavior, and calibration.

World models in driving can be divided into at least four roles. A \emph{representation world model} pretrains spatio-temporal features for downstream tasks, as in DriveWorld \cite{min2024driveworld}. A \emph{generative scene model} predicts future sensor observations or BEV states, as in GenAD and Drive-WM \cite{yang2024generalized,wang2024driving}. A \emph{trajectory evaluator} scores candidate plans by imagined consequences, as in WoTE and World4Drive \cite{li2025end,zheng2025world4drive}. A \emph{closed-loop simulator or renderer} attempts to support interactive evaluation, as in Bench2Drive-R and WorldDrive \cite{you2024bench2drive,gui2026bridging}. These roles should not be collapsed. A model that generates realistic videos is not automatically a good planner; a model that improves planning scores may not be a faithful simulator; and a simulator useful for evaluation may be too slow for onboard planning.

\begin{table*}[t]
\caption{World Models and Generative Planning.}
\label{tab:worldmodels}
\centering
\setlength{\extrarowheight}{2pt}
\begin{tabularx}{\textwidth}{p{0.16\textwidth}p{0.16\textwidth}Y Y}
\toprule
Method & World-model role & Planning connection & Open concern \\
\midrule
MILE \cite{hu2022model} & Model-based imitation & Learns latent dynamics for urban driving and connects model learning with policy learning. & Closed-loop realism depends on the learned latent dynamics and training distribution. \\
GenAD \cite{yang2024generalized} & Large-scale predictive video model & Can be adapted toward action-conditioned prediction or planning. & Video prediction quality does not directly certify safe trajectory choice. \\
DriveWorld \cite{min2024driveworld} & 4D representation pretraining & Improves downstream perception, forecasting, occupancy, and planning through spatio-temporal pretraining. & Benefit depends on transfer to planning metrics, not only representation benchmarks. \\
Drive-WM \cite{wang2024driving} & Multiview visual forecasting & Generates multiple futures and selects trajectories through image-based rewards. & Requires reliable future generation under ego interventions. \\
LAW \cite{li2025enhancing} & Latent world model & Enhances E2E driving through self-supervised future modeling. & Latent prediction errors can be hard to diagnose. \\
WoTE \cite{li2025end} & BEV world-model evaluator & Predicts future BEV states to evaluate candidate trajectories online. & Needs calibration between BEV future quality and closed-loop safety. \\
World4Drive \cite{zheng2025world4drive} & Intention-aware physical latent model & Generates, evaluates, and ranks multi-modal planning trajectories with self-supervised future alignment. & Perception-annotation-free planning still needs rare-case validation. \\
Bench2Drive-R \cite{you2024bench2drive} & Generative reactive benchmark & Uses generative rendering and behavior rollout to move real data toward reactive closed-loop evaluation. & Rendering fidelity and agent reaction fidelity must both be validated. \\
WorldDrive \cite{gui2026bridging} & Unified scene generation and planning & Shares vision and motion representations and uses future-aware reward for real-time planning. & Very recent; independent reproduction and cross-benchmark stress testing remain limited. \\
\bottomrule
\end{tabularx}
\end{table*}

For planning, the decisive evidence for a world model is not whether it generates visually realistic or statistically accurate futures in isolation, but whether it improves the selection of safe, goal-consistent, and interaction-aware ego trajectories. A world model should therefore be evaluated as a consequence estimator for candidate plans, rather than only as a generative predictor of future scenes. This requires reporting how imagined futures affect trajectory ranking, safety-progress trade-offs, rare-scenario behavior, and closed-loop outcomes under distribution shift.

\subsection{VLMs, VLA Models, and Driving Reasoning}
Vision-language models have two attractions for driving. They can encode semantic knowledge that is hard to specify in numeric labels, and they can expose decisions in a form humans can inspect. As organized in Table~\ref{tab:vla}, DriveLM's graph VQA formulation connects perception, prediction, and planning through language-structured reasoning \cite{sima2024drivelm}. LMDrive demonstrates closed-loop driving with language input \cite{shao2024lmdrive}. DriveVLM proposes a hierarchical scene description, analysis, and planning framework \cite{tian2025drivevlm}. OmniDrive adds 3D-aware multimodal reasoning for perception, planning, and counterfactual questions \cite{wang2025omnidrive}.

Recent systems move from language as annotation to language as action scaffolding. SimLingo jointly supports closed-loop driving, vision-language understanding, and language-action alignment \cite{renz2025simlingo}. ORION uses vision-language instructed action generation to bridge reasoning space and trajectory output \cite{fu2025orion}. SpaceDrive inserts spatial awareness into VLM-based driving and replaces digit-wise coordinate tokens with more task-appropriate spatial representations \cite{li2026spacedrive}. SGDrive structures VLM reasoning around a scene-agent-goal hierarchy \cite{li2026sgdrive}. WAM-Flow treats trajectory planning as discrete flow matching over structured action tokens \cite{xu2026wam}. These systems are promising because they target the core problem of aligning semantics with motion.

Still, VLA driving should be treated cautiously. Strong VLM reasoning does not automatically imply safe control. Language models may add latency, produce non-actionable rationales, or hallucinate scene facts. For deployment, it is crucial to distinguish language used for training, language used for online planning, and language used for explanation after the fact. Each role has different compute, safety, and verification implications.

\begin{table*}[t]
\caption{Language, VLM, and VLA Driving Systems.}
\label{tab:vla}
\centering
\setlength{\extrarowheight}{2pt}
\begin{tabularx}{\textwidth}{p{0.13\textwidth}p{0.18\textwidth}Y Y}
\toprule
Method & Language role & Action interface & Main limitation or open question \\
\midrule
DriveLM \cite{sima2024drivelm} & Graph VQA and planning-oriented reasoning labels. & Language supervises perception-prediction-planning questions rather than directly controlling the vehicle. & Needs faithful link between QA correctness and driving quality. \\
LMDrive \cite{shao2024lmdrive} & Language-guided closed-loop driving. & Uses language instruction as part of the driving input. & Generalization to ambiguous and unseen instructions remains difficult. \\
DriveVLM \cite{tian2025drivevlm} & Hierarchical scene description, analysis, and planning. & VLM supports planning decisions with structured reasoning. & Real-time deployment and verification are challenging. \\
OmniDrive \cite{wang2025omnidrive} & 3D-aware multimodal LLM-agent and counterfactual reasoning. & Couples 3D perception, reasoning, and planning tasks. & Benchmarking reasoning improvements against closed-loop driving is still immature. \\
EMMA \cite{hwang2024emma} & Generalist multimodal model with driving outputs represented in text. & Generates planner trajectories, objects, and road graph elements through task prompts. & Textual coordinate/action representation is expensive and can be awkward for precise control. \\
DiMA \cite{hegde2025distilling} & Distills MLLM knowledge into a vision planner. & Language model can be removed or reduced at inference. & Distillation may lose rare semantic reasoning capabilities. \\
SimLingo \cite{renz2025simlingo} & Vision-language understanding plus language-action alignment. & Supports closed-loop driving and aligns language with future actions. & Requires evaluation that penalizes explanation-action inconsistency. \\
ORION \cite{fu2025orion} & Vision-language instructed action generation. & Uses LLM reasoning and a generative planner for trajectory output. & Bridging semantic reasoning and precise trajectory generation remains delicate. \\
SpaceDrive \cite{li2026spacedrive} & Spatially aware VLM reasoning. & Uses coordinate encodings to improve trajectory regression and spatial grounding. & Depends on depth/coordinate quality and spatial calibration. \\
SGDrive \cite{li2026sgdrive} & Scene-agent-goal hierarchy for driving cognition. & Structures VLM representation toward planning. & Very recent; needs cross-benchmark and closed-loop validation. \\
WAM-Flow \cite{xu2026wam} & VLA trajectory token generation. & Parallel discrete flow matching over structured action tokens. & Needs careful reward alignment and independent reproduction. \\
\bottomrule
\end{tabularx}
\end{table*}

Similarly, the key evidence for VLA driving is not linguistic plausibility, explanation quality, or visual question-answering accuracy alone, but whether language-grounded reasoning improves trajectory choice, recovery behavior, controllability, and long-tail decision quality. A VLA planner should therefore be evaluated by action-grounded metrics: whether its language inputs, rationales, or latent semantic representations lead to safer and more route-compliant plans under benchmarked driving protocols. Without this link, language remains an auxiliary explanatory layer rather than a validated planning mechanism.

% \FloatBarrier
\section{Benchmarks and Metrics}
\subsection{From Datasets to Planning Benchmarks}
Datasets such as BDD100K, nuScenes, Waymo Open Dataset, and Waymo Open Motion Dataset remain foundational resources for perception, tracking, forecasting, and planning-related tasks \cite{yu2020bdd100k,caesar2020nuscenes,sun2020scalability,ettinger2021large}. However, a dataset is not automatically a driving benchmark. A planning benchmark must define an input interface, an output interface, a scenario distribution, metrics, and comparison rules. CARLA enabled closed-loop simulation and route-based leaderboards \cite{dosovitskiy2017carla}; nuPlan provided a large-scale ML planning benchmark with open-loop and closed-loop protocols \cite{caesar2021nuplan}; Bench2Drive standardized multi-ability closed-loop evaluation in CARLA \cite{jia2024bench2drive}; NAVSIM introduced scalable non-reactive real-log evaluation \cite{dauner2024navsim}; TAD-E2E and WOD-E2E expanded the real-world E2E data and long-tail evaluation frontier \cite{liu2025tad,xu2026wod}.

\begin{table*}[t]
\caption{Datasets and Evaluation Resources.}
\label{tab:benchmarks}
\centering
\setlength{\extrarowheight}{2pt}
\begin{tabularx}{\textwidth}{p{0.14\textwidth}p{0.18\textwidth}Y Y}
\toprule
Resource & Type & Main contribution & Planning-oriented caveat \\
\midrule
CARLA \cite{dosovitskiy2017carla} & Closed-loop simulator & Enables interactive route following, scenario testing, and leaderboards. & Simulation realism and route randomness can affect comparability. \\
nuScenes \cite{caesar2020nuscenes} & Real-world multimodal dataset & Standard perception and planning-related dataset with cameras, LiDAR, radar, maps, and annotations. & Open-loop planning on logged scenes does not guarantee closed-loop behavior. \\
nuPlan \cite{caesar2021nuplan} & Real-world planning benchmark & Introduces large-scale planning evaluation with closed-loop simulator and planning metrics. & Metric implementation and simulator settings must be reported carefully. \\
Bench2Drive \cite{jia2024bench2drive} & Closed-loop E2E benchmark & Provides standardized training data and multi-ability CARLA evaluation across interactive scenarios. & Still inherits CARLA sim-to-real limitations. \\
NAVSIM \cite{dauner2024navsim} & Non-reactive real-log benchmark & Scalable real-sensor evaluation through PDMS and pseudo-simulation. & Background agents do not react to ego behavior. \\
DriveLM \cite{sima2024drivelm} & Language/reasoning dataset & Adds graph VQA and planning-oriented language supervision. & Language correctness must be tied to action quality. \\
TAD-E2E \cite{liu2025tad} & Large-scale real-world E2E dataset & Expands scale and complexity for E2E driving data. & Requires careful protocol alignment with other benchmarks. \\
WOD-E2E \cite{xu2026wod} & Long-tail E2E benchmark & Focuses on rare scenarios and Rater Feedback Score. & Preference-aware open-loop evaluation is new and needs broader stress testing. \\
\bottomrule
\end{tabularx}
\end{table*}

\begin{table*}[t]
\caption{Evaluation Protocols and Supported Claims.}
\label{tab:protocols}
\centering
\setlength{\extrarowheight}{2pt}
\begin{tabularx}{\textwidth}{p{0.18\textwidth}p{0.17\textwidth}Y Y}
\toprule
Protocol & Examples & What it measures well & What it cannot establish alone \\
\midrule
Open-loop log replay & nuScenes planning splits, WOD-E2E trajectory prediction & Imitation fidelity, fast iteration, large-scale real-log evaluation. & Recovery, interaction, compounding errors, and whether a different safe trajectory is preferable. \\
Safety-aware open-loop or pseudo-simulation & NAVSIM, NAVSIM-v2 style metrics & Scalable planning proxies using real sensor inputs and safety/progress/comfort submetrics. & Full reactivity of other agents and closed-loop recovery after ego-induced deviations. \\
Reactive closed-loop simulation & CARLA Leaderboard, Bench2Drive, nuPlan closed-loop & Interaction, route completion, infractions, comfort, and recovery under the policy's own state distribution. & Sim-to-real validity, rare real-world semantics, and human preference in ambiguous scenes. \\
Generative reactive evaluation & Bench2Drive-R, emerging world-model simulators & Potential bridge between real data and reactive closed-loop testing. & Rendering and behavior model fidelity must be validated separately. \\
Long-tail stress tests & WOD-E2E, Fail2Drive, hazardous scenario generation & Rare events, distribution shift, paired generalization drops, and preference-sensitive decisions. & Average driving competence across normal routes. \\
Cross-benchmark correlation & NAVSIM--Bench2Drive correlation studies \cite{wang2026open} & Whether proxy metrics predict closed-loop ranking. & True safety; correlations can be sample-dependent and non-monotonic. \\
Real-world closed-loop testing & Industrial robotaxi or ADAS deployments & Actual operational behavior and safety monitoring. & Public reproducibility, controlled ablations, and fair academic comparison. \\
\bottomrule
\end{tabularx}
\end{table*}

\subsection{Metric Families}
Open-loop metrics include average displacement error (ADE), final displacement error (FDE), trajectory L2 error, and collision proxies computed against logged scenes. They are useful for fast iteration but weak as final evidence. A model may match the logged trajectory without being able to recover from deviations, and a different trajectory may be safer or more natural than the logged human path. The paper ``Is Ego Status All You Need?'' showed that open-loop protocols can be vulnerable to shortcuts in which models rely heavily on ego-state signals rather than perception \cite{li2024ego}. Hidden Biases similarly demonstrated that benchmark protocols can distort conclusions about E2E driving models \cite{jaeger2023hidden}.

Closed-loop metrics evaluate what happens when the policy acts. As summarized in Table~\ref{tab:benchmarks}, CARLA Leaderboard and Bench2Drive report driving score, route completion, infraction penalties, success rate, efficiency, and comfort-related quantities \cite{jia2024bench2drive}. nuPlan combines safety, progress, drivable-area compliance, time-to-collision, comfort, and traffic-rule indicators \cite{caesar2021nuplan}. NAVSIM's Predictive Driver Model Score (PDMS) offers a scalable middle ground by evaluating planned trajectories in a non-reactive real-log setting \cite{dauner2024navsim}. WOD-E2E introduces Rater Feedback Score (RFS), which measures alignment with rater-preferred trajectory labels in long-tail scenarios \cite{xu2026wod}.

\subsection{Why Open-Loop and Closed-Loop Disagree}
\begin{figure*}[t]
\centering
\includegraphics[width=0.90\textwidth]{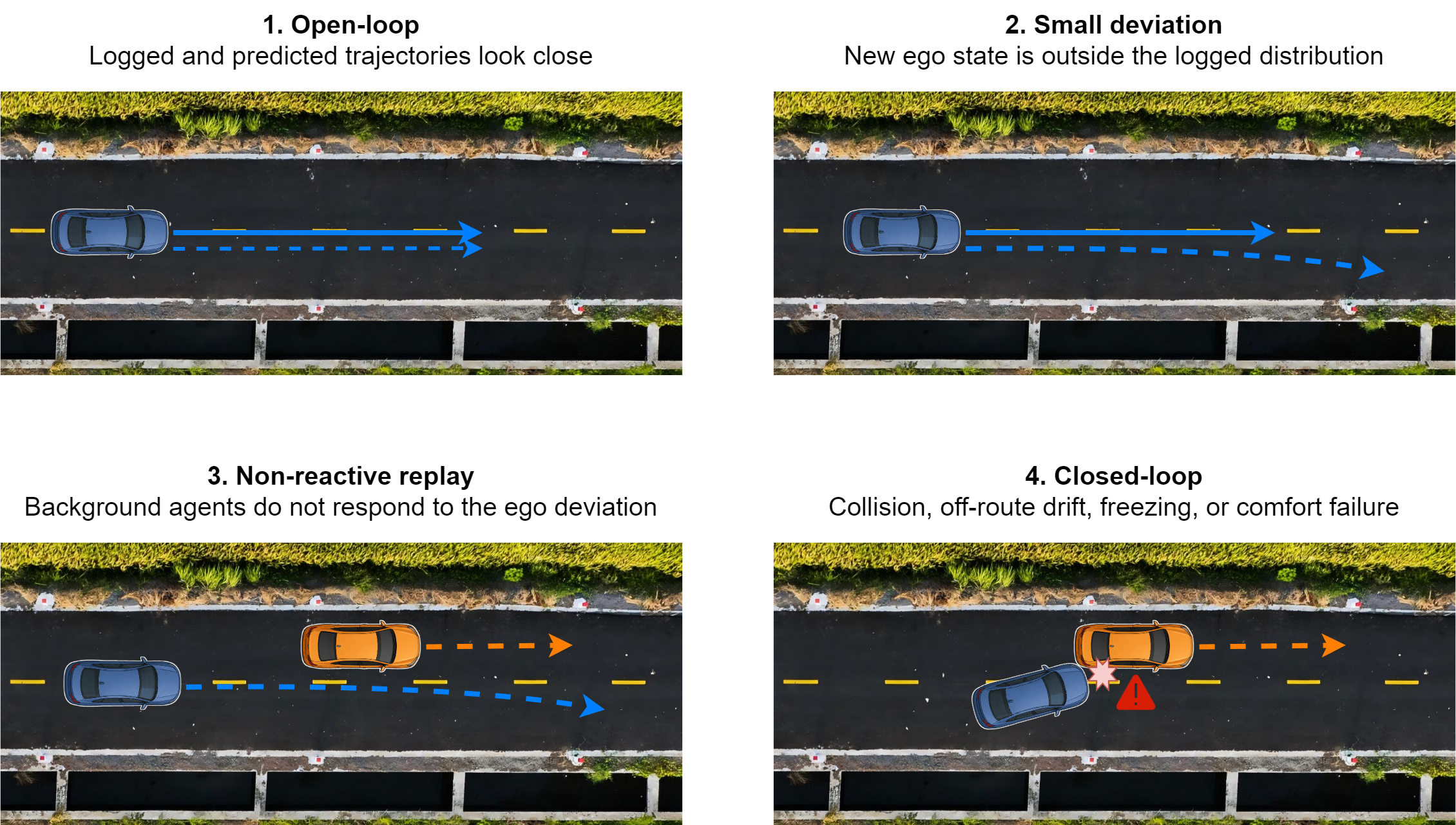}
\caption{Why open-loop and closed-loop evaluations can disagree. Open-loop trajectory matching tests similarity under the logged state distribution, while closed-loop evaluation tests the policy under the states it actually induces.}
\label{fig:metric_failure}
\end{figure*}

Open-loop and closed-loop metrics disagree for structural reasons as illustrated in Figure~\ref{fig:metric_failure}. First, open-loop evaluation scores a policy under the expert's state distribution, while closed-loop evaluation scores it under its own induced state distribution. Second, open-loop metrics often penalize valid alternatives, especially in multimodal interactions. Third, open-loop metrics typically lack feedback from background agents and rule violations. Fourth, closed-loop systems depend on controller details, simulator dynamics, and stochastic scenario outcomes that may not appear in open-loop logs.

This disagreement has practical consequences for survey writing. A method should not be called state of the art without specifying the protocol. A strong nuScenes L2 score, a strong NAVSIM PDMS score, and a strong Bench2Drive Driving Score are different claims. Comparative tables are more reliable when they group methods by benchmark and metric family rather than mixing scores across protocols.

Recent correlation work makes this point sharper. Cross-referencing NAVSIM and Bench2Drive results suggests that safety-aware open-loop scores can correlate with closed-loop driving score better than ADE or FDE, but ranking inversions remain and submetrics can saturate \cite{wang2026open}. NAVSIM-style scores are valuable because they are scalable and more planning-aware than L2, but they should be described as proxy evidence, not as a substitute for closed-loop interaction. Conversely, Bench2Drive-style scores are more behaviorally meaningful, but they remain tied to CARLA's simulator distribution. A robust empirical claim should therefore triangulate across at least one real-log proxy and one reactive closed-loop protocol.

\subsection{Recommended Reporting Protocol}
For future E2E-AD papers, we recommend reporting five groups of evidence: open-loop trajectory quality, non-reactive real-log planning score, closed-loop route and infraction metrics, long-tail or preference-aware performance, and qualitative failure analysis. Authors should also report sensor configuration, route inputs, output type, controller, safety wrapper, training data source, data filtering, compute budget, random seeds, benchmark version, and metric implementation. Without these details, small leaderboard differences are difficult to interpret.

Following Table \ref{tab:protocols}, we recommend avoiding a single ``best method'' column. A more honest format is to use separate columns for open-loop imitation, proxy planning, reactive closed-loop, long-tail shift, code availability, and compute. A method that is weak in open-loop L2 but strong in closed-loop route completion may be a better planner. A method that is strong on NAVSIM but slow in Bench2Drive may be over-optimizing safety at the expense of progress. A VLA method that produces strong rationales but mediocre trajectories should not be promoted as a driving breakthrough. The unit of comparison should be the claim, not the model name.

% \FloatBarrier
\section{Safety, Robustness, and Deployability}
\subsection{From Black-Box Critique to Auditable Structure}
E2E systems are often criticized as black boxes. The stronger version of this critique is not that all learned systems are uninterpretable, but that their failure modes are difficult to localize when the entire stack is optimized jointly. Recent work addresses this through interpretable intermediate structure, causal analysis, and safety constraints. Safety-critical evaluation settings are summarized in Table~\ref{tab:safety}. InterFuser uses interpretable sensor fusion and safety-enhanced action constraints \cite{shao2023safety}. Hidden Biases exposes protocol pitfalls \cite{jaeger2023hidden}. Causal analysis of E2E driving introduces interventions and counterfactual reasoning into model debugging \cite{li2024exploring}. DriveLM and related language systems attempt to expose scene reasoning, but explanation alone is insufficient unless it is faithful to the action output \cite{sima2024drivelm}.

\subsection{Safety Guarantees Remain Limited}
Most public E2E driving papers provide empirical safety improvements rather than formal guarantees. These improvements include better benchmark metrics, collision-aware losses, occupancy constraints, route filters, emergency braking modules, uncertainty-aware planners, and preference-aware evaluation. Formal verification of high-dimensional E2E driving policies remains much less developed than empirical benchmarking. This gap should be stated plainly. Planning-oriented E2E systems can be safer than earlier direct-control policies, but public evidence is still dominated by benchmark performance rather than deployable safety certificates.

\begin{table*}[t]
\caption{Safety and Interpretability Mechanisms.}
\label{tab:safety}
\centering
\setlength{\extrarowheight}{2pt}
\begin{tabularx}{\textwidth}{p{0.18\textwidth}Y Y Y}
\toprule
Mechanism & How it is used & Strength & Limitation \\
\midrule
Structured intermediate features & BEV, occupancy, vectors, object tokens, or map elements remain visible to the planner. & Supports debugging and compatibility with rule checks. & Intermediate correctness does not guarantee final action safety. \\
Safety-aware losses & Penalize collisions, off-road motion, traffic-rule violations, or uncomfortable acceleration. & Easy to train and benchmark. & Loss weights can overfit to metrics or produce overly conservative behavior. \\
Runtime filters & Emergency braking, drivable-area filters, occupancy constraints, or fallback policies. & Provides a last line of empirical defense. & Can mask planner weaknesses and may cause deadlock or abrupt behavior. \\
Uncertainty estimates & Probabilistic plans, diffusion samples, ensemble uncertainty, or calibrated risk scores. & Helps identify ambiguous scenes and multi-modal futures. & Uncertainty can be poorly calibrated under distribution shift. \\
Causal diagnostics & Counterfactual scene edits and interventions \cite{li2024exploring}. & Reveals shortcut learning and failure causes. & Expensive to scale; counterfactual validity is hard. \\
Language rationales & DriveLM, LMDrive, Drive-R1, EMMA, SimLingo-style explanations \cite{sima2024drivelm,shao2024lmdrive,hwang2024emma,renz2025simlingo,li2026drive}. & Improves human inspection and instruction following. & Explanations may not be faithful to the trajectory generator. \\
Long-tail benchmarks & WOD-E2E and Fail2Drive stress rare scenarios and shifts \cite{xu2026wod,gerstenecker2026fail2drive}. & Exposes failures hidden by average scores. & Coverage of real-world rare events is still incomplete. \\
\bottomrule
\end{tabularx}
\end{table*}

\subsection{Runtime Assurance and Safety Envelopes}
Empirical benchmark safety is not the same as operational safety assurance. A practical E2E driving stack will likely require runtime mechanisms that constrain or monitor the learned planner. These mechanisms can be added without abandoning end-to-end learning: the neural policy proposes trajectories, while an assurance layer checks whether the proposal remains inside a verified or rule-constrained envelope. This view is compatible with planning-oriented E2E-AD because it treats learning as a source of candidate behavior, not as the only safety mechanism in the vehicle.

Several engineering routes are relevant. A safety monitor can reject plans that violate drivable area, speed limit, route topology, or time-to-collision thresholds. A fallback planner can switch to conservative braking or rule-based lane keeping when confidence is low. Rule-based shields can enforce traffic-light, stop-sign, or lane-boundary constraints. Control barrier functions and reachability analysis can define safe sets for low-level control, though scaling them to perception-conditioned, multi-agent, high-dimensional E2E policies remains difficult. Emergency braking modules can reduce collision risk but may create comfort and rear-end risks if triggered too often. Uncertainty thresholding, out-of-distribution detection, and calibration can decide when the learned policy should defer to a safer mode.

The key research question is where the assurance boundary should sit. If the shield operates only after trajectory generation, it may prevent collisions but not improve upstream reasoning. If it is differentiable and included during training, it can shape the planner but may be harder to certify. If the monitor is too conservative, it can create freezing and low-progress behavior; if it is too permissive, it becomes a weak post-hoc filter. Future E2E papers should therefore report not only whether a safety wrapper is used, but also its trigger conditions, intervention frequency, and effect on progress, comfort, and infractions.

\subsection{Compute, Latency, and Deployability}
Unified stacks, world models, and VLA systems can be expensive. Deployment imposes latency, memory, power, and verification constraints that are often underreported in benchmark papers. PARA-Drive explicitly targets real-time operation \cite{weng2024drive}. DiMA addresses the cost of MLLM reasoning through distillation \cite{hegde2025distilling}. DrivoR compresses multi-camera features with register tokens \cite{kirby2026driving}. WAM-Flow explores parallel trajectory decoding with a tunable compute-accuracy trade-off \cite{xu2026wam}. These systems point toward a likely future: E2E driving research will increasingly be judged not only by accuracy but also by whether the reasoning stack can run within realistic compute budgets.

Compute should be treated as a first-class scientific variable. A large VLM planner can look compelling if evaluated offline, but a vehicle needs bounded latency, predictable memory use, and robust behavior when the model is uncertain. Distillation, register tokens, sparse queries, and parallel decoding are therefore not merely engineering optimizations; they determine whether reasoning-rich E2E systems can be evaluated fairly and deployed plausibly. For this reason, survey tables should record model size, sensor history, frame rate, online language-model usage, and whether the method uses test-time sampling or ensembling.

% \FloatBarrier
\section{Reproducibility and Future Directions}
Beyond benchmark scores, the long-term value of planning-oriented E2E-AD depends on whether methods can be reproduced, compared, and stress-tested under aligned protocols. This section summarizes reporting requirements, public-resource levels, and open challenges for future research.

\subsection{Reproducibility Checklist}
Table~\ref{tab:checklist} lists items that should be extracted from papers whenever possible. If a paper does not report an item, it should be marked as unavailable.

\begin{table*}[t]
\caption{Reproducibility Checklist.}
\label{tab:checklist}
\centering
\setlength{\extrarowheight}{2pt}
\begin{tabularx}{\textwidth}{p{0.18\textwidth}Y Y}
\toprule
Item & Minimum reporting requirement & Why it matters \\
\midrule
Benchmark version & Dataset split, simulator version, metric version, leaderboard date, and official-server versus local evaluation. & Avoids hidden protocol mismatch and stale leaderboard comparisons. \\
Sensor setup & Camera count, resolution, history length, LiDAR/radar use, ego state, map inputs, and route or command interface. & Determines whether two methods solve the same observation problem. \\
Output and controller & Direct control, waypoint sequence, deterministic or probabilistic trajectory, action tokens, and PID/MPC/learned controller details. & Separates planner quality from controller effects and post-processing. \\
Training signal & Human BC, simulator expert, privileged distillation, auxiliary tasks, RL, world-model loss, language QA, preference labels, or feedback correction. & Explains whether gains come from architecture, supervision, or expert design. \\
Data curation & Filtering, hard-scenario mining, perturbations, augmentation, pseudo-expert generation, and long-tail sampling. & Makes robustness claims auditable rather than anecdotal. \\
Runtime structure & Safety filter, emergency brake, drivable-area constraint, uncertainty threshold, fallback policy, or post-planning optimizer. & Clarifies whether reported safety comes from the learned planner or external assurance. \\
Compute budget & GPU type, GPU count, training time, inference latency, model size, sampling count, and online language-model use. & Enables deployment-aware and compute-normalized comparison. \\
Statistics & Number of runs, random seeds, confidence intervals, route variance, and failure-case buckets. & Distinguishes stable improvements from benchmark noise or cherry-picked routes. \\
\bottomrule
\end{tabularx}
\end{table*}

\subsection{Open-Source Ecosystem}
Not all public resources support the same level of reproducibility. A paper may release inference code but not training code; a benchmark may release data and metrics but require a hidden server for final comparison; a method may release a checkpoint that works only under a specific simulator version. It is useful to distinguish at least four reproducibility levels: \emph{paper-only}, where only results are reported; \emph{inference-reproducible}, where code or checkpoints allow evaluation; \emph{training-reproducible}, where scripts, data processing, and configurations allow retraining; and \emph{benchmark-reproducible}, where official metrics, versions, and evaluation containers are stable enough for independent comparison. Table~\ref{tab:opensource} summarizes the role of several public resources.

\begin{table*}[t]
\caption{Public Resources and Reproducibility Levels.}
\label{tab:opensource}
\centering
\setlength{\extrarowheight}{2pt}
\begin{tabularx}{\textwidth}{p{0.16\textwidth}p{0.18\textwidth}Y Y}
\toprule
Resource & Reproducibility role & Useful for & Main caution \\
\midrule
CARLA \cite{dosovitskiy2017carla} & Simulator platform & Closed-loop route following, scenario design, expert data generation. & Simulator version, route set, traffic manager, and random seed can change results. \\
TransFuser/TCP lineage \cite{prakash2021multi,wu2022trajectory} & Public baselines and strong CARLA agents & Comparing sensor fusion, trajectory-control coupling, and closed-loop behavior. & Results are sensitive to route protocol and controller details. \\
UniAD \cite{hu2023planning} & Unified planning-oriented stack & Studying perception-prediction-planning coupling and multi-task losses. & Heavy training and nuScenes-centered evaluation can limit easy replication. \\
VAD/VADv2 \cite{jiang2023vad,jiang2024vadv2} & Vectorized and probabilistic planning baselines & Testing vector representation and uncertainty-aware planning. & Open-loop performance needs closed-loop triangulation. \\
Bench2Drive \cite{jia2024bench2drive} & Standardized closed-loop E2E benchmark & Multi-ability evaluation, fixed training data, CARLA closed-loop comparison. & Sim-to-real gap and leaderboard variance remain. \\
NAVSIM \cite{dauner2024navsim} & Scalable non-reactive real-log benchmark & Fast proxy evaluation with safety/progress/comfort metrics. & Background agents are not reactive; ranking inversions with closed-loop can occur. \\
SimLingo \cite{renz2025simlingo} & Language-action aligned closed-loop agent & Studying whether language helps both explanation and control. & Language tasks and driving tasks must be evaluated jointly. \\
LEAD \cite{nguyen2026lead} & Student-aware expert and dataset pipeline & Testing learner-expert asymmetry and improved imitation data generation. & Benefits should be separated from architecture and data-scale effects. \\
WOD-E2E \cite{xu2026wod} & Long-tail real-world E2E benchmark & Rater Feedback Score, rare scenario evaluation, preference-aware planning. & Still open-loop; preference labels and hidden test sets limit full local reproduction. \\
Fail2Drive \cite{gerstenecker2026fail2drive} & Paired closed-loop generalization benchmark & Measuring performance drop under controlled distribution shifts. & CARLA-based shifts are diagnostic, not exhaustive real-world coverage. \\
\bottomrule
\end{tabularx}
\end{table*}

\begin{table*}[t]
\caption{Open Problems and Needed Evidence.}
\label{tab:openproblems}
\centering
\setlength{\extrarowheight}{2pt}
\begin{tabularx}{\textwidth}{p{0.18\textwidth}Y Y Y}
\toprule
Problem & Why it is unsolved & Promising direction & What evidence is needed \\
\midrule
Metric validity & Open-loop, non-reactive, closed-loop, and preference metrics test different properties. & Cross-benchmark correlation and causal analysis of metric failures. & Paired evaluations across NAVSIM, Bench2Drive, WOD-E2E, and real-world or high-fidelity simulation. \\
Long-tail robustness & Rare events are too sparse for ordinary imitation and too diverse for fixed scenario suites. & Data mining, preference labels, hazardous scenario generation, and paired generalization tests. & Degradation curves, not just average scores. \\
Learner-expert asymmetry & Expert demonstrations may use unobservable or unrealistically precise state. & Student-aware expert design, multi-teacher distillation, uncertainty-aware imitation. & Controlled experiments holding architecture and data scale fixed. \\
World-model calibration & Generated futures may look plausible but fail under intervention. & Uncertainty-aware world models, action-conditioned rollouts, future-aware rewarders. & Calibration, rare-case prediction, and closed-loop plan-selection tests. \\
Language-action grounding & VLMs reason in semantic space while vehicles execute metric trajectories. & Spatial tokens, action-aligned language supervision, distilled planners, structured VLA decoders. & Ablations showing trajectory improvement, not only better explanations. \\
Formal and empirical safety & Public E2E work mostly reports benchmark mitigation rather than guarantees. & Runtime monitors, verified envelopes, uncertainty thresholds, fallback planning. & Failure-rate confidence intervals, monitor recall, and adversarial or counterfactual testing. \\
Compute and deployment & Large VLM/world-model systems may be too slow or expensive for onboard use. & Sparse queries, register tokens, distillation, parallel decoding, adaptive compute. & Latency-normalized and energy-normalized benchmark reports. \\
Open reproducibility & Code, data, checkpoints, and exact metric versions are unevenly released. & Standardized reproducibility cards and official evaluation containers. & Independent reruns and public failure-case logs. \\
\bottomrule
\end{tabularx}
\end{table*}

This distinction also changes how comparative claims should be written. A method with paper-only results should not be weighted the same as a method with code, checkpoints, and official benchmark entries. A benchmark with stable metrics should be treated differently from a one-off evaluation script. For a publishable survey, each method should be marked by code availability, checkpoint availability, training reproducibility, evaluation server, and license whenever such information is publicly available. In E2E-AD, small differences in sensor preprocessing, route command encoding, controller parameters, or metric thresholds can produce large differences in reported driving score.

Reproducibility also affects citation value. Readers often cite surveys not only for conceptual taxonomy, but also for tables that help them choose baselines. A survey that marks which baselines are actually runnable will be more useful than one that only lists SOTA claims. This is especially important for world-model and VLA papers, where computation, data access, and inference-time language model dependencies can make nominal reproduction unrealistic. For frontier methods, the survey should therefore separate \emph{conceptual importance} from \emph{ready-to-use baseline status}.

\subsection{Open Challenges and Future Directions}
Table \ref{tab:openproblems} shows some open problems and possible solutions that still exist in current E2E-AD research. Future benchmark-centered works could add small empirical extensions, such as cross-protocol correlation, representation ablations, language-role ablations, world-model evaluator tests, and compute-normalized comparisons. These extensions should be framed as reproducibility studies rather than new unofficial leaderboards: the goal is to test whether qualitative conclusions survive across open-loop, non-reactive, closed-loop, and long-tail protocols.

\subsubsection{Benchmark Unification}
The field still lacks a single protocol that combines real-world sensor realism, reactive interaction, long-tail coverage, scalable evaluation, human preference, and standardized metrics. nuPlan, Bench2Drive, NAVSIM, TAD-E2E, and WOD-E2E each solve part of the problem \cite{caesar2021nuplan,jia2024bench2drive,dauner2024navsim,liu2025tad,xu2026wod}. A major future direction is to report cross-protocol performance and study which metrics predict true closed-loop safety and route success.

\subsubsection{Uncertainty-Aware and Preference-Aware Planning}
Deterministic trajectory regression is poorly matched to multimodal driving futures. Probabilistic planning, diffusion, flow matching, and world-model candidate evaluation are promising responses \cite{jiang2024vadv2,liao2025diffusiondrive,li2025end,xu2026wam}. Preference-aware evaluation, as in WOD-E2E, adds another layer: the best plan may not be the closest plan to a logged trajectory, but the one judged safer or more appropriate by human raters \cite{xu2026wod}. Future systems should learn from risk, rules, comfort, and preference feedback rather than treating these only as test metrics.

\subsubsection{Causal and Counterfactual Evaluation}
E2E models can exploit shortcuts that are invisible under aggregate scores. Causal interventions, counterfactual scene edits, and targeted stress tests are needed to determine whether a policy uses relevant visual evidence, understands agent interaction, and respects route intent \cite{li2024exploring,jaeger2023hidden}. This is especially important for VLA systems, where language explanations can conceal rather than reveal the true action mechanism.

\subsubsection{Language-Action Grounding}
The next generation of VLA driving models must prove that language improves action quality. This requires benchmarks where language, scene understanding, and trajectories are jointly evaluated. It also requires separating training-time language supervision from inference-time language reasoning. A deployed vehicle may benefit from distilled reasoning without running a large language model in the control loop \cite{hegde2025distilling}. Conversely, some long-tail scenarios may require online semantic reasoning, especially when the scene contains unusual agents, ambiguous human signals, or instructions not captured by route geometry.

\subsubsection{World-Model Reliability}
World models are attractive because they can evaluate consequences, but inaccurate imagined futures can be dangerous. Future work should report calibration, failure modes, uncertainty, and rare-event performance for world-model components. A planning system should know when its learned future model is unreliable. This may require hybrid designs that combine latent imagination with explicit constraints, simulator checks, and conservative fallback policies.

\subsubsection{Public Baselines and Reproducible Stacks}
The field needs strong, maintained, reproducible public baselines. Bench2Drive, NAVSIM, LEAD, SimLingo, and related open stacks are valuable because they make comparison easier \cite{jia2024bench2drive,dauner2024navsim,nguyen2026lead,renz2025simlingo}. Compared to studies that only report isolated scores, a study can increase its value by publishing open-source code, public checkpoints, official benchmark entries, and reproducible training scripts.

\subsection{Limitations of This Survey}
This survey is necessarily biased toward public academic and industrial research that exposes enough detail to be cited and compared. Proprietary L4 autonomy systems may use planning-oriented learned components, world models, or VLMs internally without publishing comparable architectures or metrics. As a result, public literature may overrepresent benchmark-driven designs and underrepresent safety validation infrastructure, fleet monitoring, simulation at industrial scale, and operational design domain management.

A second threat is recency. Several 2025--2026 papers are important because they define emerging directions, but their independent reproduction and long-term citation impact are not yet known. This survey therefore treats very recent work such as WOD-E2E, SpaceDrive, SGDrive, Fail2Drive, and WorldDrive as frontier signals rather than settled consensus. A third threat is metric dependence. Many claims in E2E-AD are only meaningful within a specific benchmark version, sensor setup, and controller. The survey's comparisons should therefore be read as a structured map of claims, not as a universal ranking of methods.

\section{Conclusion}
End-to-end autonomous driving has matured from direct-control imitation into a family of planning-oriented systems built on structured representations, richer supervision, world-model reasoning, language-action alignment, and more serious evaluation protocols. Across the reviewed literature, end-to-end optimization increasingly means learning representations and policies whose structure serves the final planning objective, rather than eliminating structure altogether. As a result, the scientific questions have changed. The field no longer asks only whether pixels can be mapped to controls. It asks which representations support safe planning, which supervision signals produce robust closed-loop behavior, which metrics reflect actual driving quality, and how uncertainty, language, and imagined futures should be integrated without sacrificing reliability.

The evaluation shift is one of the most visible recent changes. Open-loop trajectory errors are useful diagnostics, but they are not sufficient evidence for driving competence. Planning-oriented E2E-AD is better assessed under benchmark-consistent protocols that measure safety, progress, comfort, route compliance, controllability, long-tail robustness, and human preference. The central thesis of this survey is therefore that planning-oriented E2E-AD should not be judged by how end-to-end it appears architecturally, but by whether its learned structure, supervision, and evaluation protocol jointly support safe and reproducible planning. In this sense, the next phase of E2E driving research is not merely architectural. It is also methodological: the field must make its claims comparable, its failures auditable, and its planning objectives faithful to the complexity of real driving.

\section*{Acknowledgment}

This work was supported by the Science and Technology Development Fund of Macau [0007/2025/RIC, 0122/2024/RIB2, 0215/2024/AGJ, 0074/2025/AMJ, 001/2024/SKL, 0002/2025/EQP], the Research Services and Knowledge Transfer Office, University of Macau [SRG2023-00037-IOTSC, MYRG-GRG2024-00284-IOTSC], the Shenzhen-Hong Kong-Macau Science and Technology Program Category C [SGDX20230821095159012], the Science and Technology Planning Project of Guangdong [2025A0505010016], National Natural Science Foundation of China [52572354], the State Key Lab of Intelligent Transportation System [2024-B001], and the Jiangsu Provincial Science and Technology Program [BZ2024055].

% \FloatBarrier

\bibliographystyle{IEEEtran}
\bibliography{references}

\end{document}